\documentclass[lettersize,journal]{IEEEtran}
\usepackage{amsmath,amsfonts}
\usepackage{algorithmic}
\usepackage{algorithm}
\usepackage{array}
\usepackage[caption=false,font=normalsize,labelfont=sf,textfont=sf]{subfig}
\usepackage{textcomp}
\usepackage{stfloats}
\usepackage{url}
\usepackage{verbatim}
\usepackage{graphicx}
\usepackage{multirow}
\usepackage{amssymb}
\usepackage{amsthm}
\usepackage{booktabs}
\usepackage[table]{xcolor}
\usepackage[numbers]{natbib}
\usepackage[pagebackref,breaklinks,colorlinks]{hyperref}
\begin{document}

\title{Wholly-WOOD: Wholly Leveraging Diversified-quality Labels for Weakly-supervised Oriented Object Detection}

\author{Yi Yu,
Xue Yang,
Yansheng Li,~\IEEEmembership{Senior Member,~IEEE},
Zhenjun Han\\
Feipeng Da,
Junchi Yan,~\IEEEmembership{Senior Member,~IEEE}
\thanks{Yi Yu and Xue Yang contribute equally to this work. Corresponding author: Junchi Yan.}
\thanks{Yi Yu and Feipeng Da are with School of Automation, Southeast University, Nanjing, 210096, China (e-mail: yuyi@seu.edu.cn; dafp@seu.edu.cn).}
\thanks{Xue Yang is with Department of Automation, Shanghai Jiao Tong University, Shanghai, 200240, China (e-mail: yangxue-2019-sjtu@sjtu.edu.cn).}
\thanks{Junchi Yan is with School of Artificial Intelligence, Shanghai Jiao Tong University, Shanghai, 200240, China (e-mail: yanjunchi@sjtu.edu.cn).}
\thanks{Yansheng Li is with School of Remote Sensing and Information Engineering, Wuhan University, Wuhan, 430079, China (e-mail: yansheng.li@whu.edu.cn).}
\thanks{Zhenjun Han is with School of Electronic, Electrical and Communication Engineering, University of Chinese Academy of Science, Beijing, 100049, China (e-mail: hanzhj@ucas.ac.cn).}
}

\markboth{Journal of \LaTeX\ Class Files,~Vol.~14, No.~8, August~2021}%
{Shell \MakeLowercase{\textit{et al.}}: A Sample Article Using IEEEtran.cls for IEEE Journals}

\IEEEpubid{0000--0000/00\$00.00~\copyright~2021 IEEE}

\maketitle

\begin{abstract}
Accurately estimating the orientation of visual objects with compact rotated bounding boxes (RBoxes) has become a prominent demand, which challenges existing object detection paradigms that only use horizontal bounding boxes (HBoxes). To equip the detectors with orientation awareness, supervised regression/classification modules have been introduced at the high cost of rotation annotation. Meanwhile, some existing datasets with oriented objects are already annotated with horizontal boxes or even single points. It becomes attractive yet remains open for effectively utilizing weaker single point and horizontal annotations to train an oriented object detector (OOD). We develop Wholly-WOOD, a weakly-supervised OOD framework, capable of wholly leveraging various labeling forms (Points, HBoxes, RBoxes, and their combination) in a unified fashion. By only using HBox for training, our Wholly-WOOD achieves performance very close to that of the RBox-trained counterpart on remote sensing and other areas, significantly reducing the tedious efforts on labor-intensive annotation for oriented objects. 
\end{abstract}


\section{Introduction}\label{sec:intro}
\IEEEPARstart{I}{n} modern computer vision applications, oriented object detection has emerged as an essential bridge to close the gap between the limited orientation resolution of traditional object detectors (i.e. based on horizontal bounding box) and the increasing demand for fine-grained pose estimation of visual objects. From the expansive vistas of remote sensing \cite{Yang2018Automatic, Xia2018DOTA, gui2024remote,Liu2017HRSC,zhou2024diffdet4sar} to the intricate worlds under a microscope \cite{fan2023deep,yang2024rotatedstomatanet,gong2023oriented}, and even within the dynamic environments of autonomous driving \cite{feng2021deep}, robotic grasping \cite{cheng2021grasp,holomjova2022exploring}, medical image \cite{li2019detecting}, scene text \cite{Ma2018Arbitrary, Liao2018Rotation,Zhou2017EAST}, retail scenes \cite{pan2020dynamic}, manufacturing \cite{li2024pcbssd}, agriculture \cite{liu2022wsrd, zhao2022deep, song2022precise}, face detection~\cite{shi2018real}, power grid equipment \cite{wei2022detecting,lin2023automatic}, insect detection \cite{nedeljkovic2023yudo}, and transverse aeolian ridges of Mars \cite{cao2024aitars}, its impact resonates across industries. Diverging from traditional detection~\cite{liu2020deep}, oriented detection introduces rotated bounding boxes that align with the orientation of objects, thereby capturing a more precise depiction. This level of detail is crucial for various applications, especially in predicting the relationships between objects within a scene graph \cite{li2024star}, making oriented detection a burgeoning field of research \cite{Yang2022Arbitrary, yang2023detecting, zhang2023remote, Zhang2023Efficient, wen2023comprehensive, xiao2024theoretically}. 

To teach the detector new concepts of visual objects, a common way is to use manual annotations. In general, objects can be annotated in four different ways: single point (Point), horizontal bounding box (HBox), rotated bounding box (RBox), and pixel-wise label (Mask). Early research typically relies on full supervision, where the manual annotation matches the desired network output format \cite{Fu2020Rotation, Ding2018Learning, Xie2021Oriented, Han2021Redet}. However, in the context of oriented detection, this approach to acquiring training data is both labor-intensive and error-prone. A fundamental contributing factor to this issue is the time-consuming nature of rotated box annotation and the vast amounts of data, especially in remote sensing \cite{Xia2018DOTA}. In concrete terms, the cost of each RBox is about 36.5\% higher than an HBox and 104.8\% higher than a point annotation\footnote{According to \url{https://cloud.google.com/ai-platform/data-labeling/pricing} and ``point annotations are 1.1-1.2$\times$ more time-consuming than obtaining image-level labels" \cite{bearman2016point}.\vspace{12pt}}. Moreover, many remote sensing images have already been annotated with HBoxes (e.g. DIOR \cite{li2020object} and SARDet-100K \cite{li2024sardet100k}). When another format is needed, re-annotation is a possible solution. For example, the aerial image dataset DIOR \citep{li2020object} has been re-annotated to build a rotated box version DIOR-RBox \cite{cheng2022anchor}, which is repetitive and inefficient.

\begin{figure*}[t]
\centering
\includegraphics[width=0.95\textwidth]{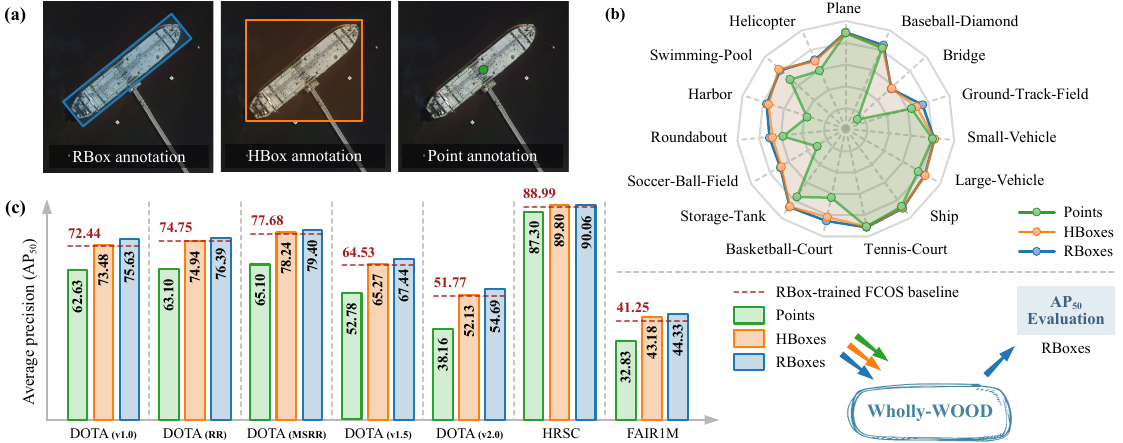}
\caption{To illustrate the task we aim at and the results we achieve. \textbf{(a)} Different annotating formats supported by our Wholly-WOOD. \textbf{(b)} Accuracy for each category of remote sensing objects in the DOTA-v1.0 dataset. \textbf{(c)} Green/Orange/Blue bars: the accuracy of Wholly-WOOD using Point/HBox/RBox annotations. Red dashed lines: the accuracy of RBox-trained FCOS \cite{Tian2019FCOS} serving as a reference to measure the accuracy disparity. }\label{fig:intro}
\end{figure*}

Such a situation raises an interesting question: Is it possible to convert annotations between different formats and make full use of available labeled data? The conversion from Mask$\rightarrow$RBox$\rightarrow$HBox$\rightarrow$Point can be easily achieved (e.g. by finding the circumscribed rectangle), while the inverse process is much more difficult, where we need to grab some additional clues from the image. Learning fine-grained labels from coarse-grained ones is usually termed weak supervision.

Research toward weakly-supervised oriented object detectors has made some progress, with several HBox-to-Mask, Point-to-Mask, and HBox-to-RBox methods being proposed (detailed in Sec. \ref{sec:work}). Particularly, the foundation model SAM (Segment Anything Model) \cite{kirillov2023segany} has shown strong zero-shot capabilities for producing object masks from the input Point/HBox prompts. Since Mask can be converted to RBox by finding circumscribed rectangles, Point/HBox-to-Mask methods (e.g. SAM) are potentially applicable to RBox generation and are compared in our experiments.

However, existing methods still exhibit shortcomings in three folds: 
\textbf{1)} While approaches like SAM \cite{kirillov2023segany} yield a zero-shot conversion, they are pre-trained on massive amounts of labeled data, negating the goal of minimizing manual labeling.
\textbf{2)} Most methods are tailored to a specific conversion, without the ability to uniformly integrate and utilize annotations when Point/HBox/RBox formats coexist.
\textbf{3)} We show that existing methods still have much room for performance improvement when compared to RBox-supervised counterparts.

There comes the motivation of our work: Oriented objects (e.g. elongated or symmetric) widely present in remote sensing and other vision task \cite{funk2017detecting}. To reduce repetitive and labor-intensive annotation work, we are intended to develop a weakly-supervised detector, capable of handling various labeled formats (i.e. Points, HBoxes, RBoxes) in a unified manner and generating RBoxes as the output.

The preliminary version of this article has partly appeared in recent conferences, including H2RBox-v2 (NeurIPS 2023) \cite{yu2023h2rboxv2} and Point2RBox (CVPR 2024) \cite{yu2024point2rbox}, where several basic principles have been devised for weakly-supervised learning. Specifically, H2RBox-v2 achieves HBox-to-RBox through symmetry-aware learning and Point2RBox achieves Point-to-RBox through synthetic pattern knowledge combination.

In this paper, we introduce Wholly-WOOD, a weakly-supervised detector for oriented object detection that unifies various label types within a single framework. The overall contributions of this extended journal version can be summarized as: 
\textbf{1)} We introduce Wholly-WOOD, a unified weakly-supervised detector for oriented objects, accommodating multiple annotation formats including Point/HBox/RBox or their combination as inputs, and producing RBox annotations as outputs\footnote{This journal version significantly extends the preliminary conference versions \cite{yu2023h2rboxv2,yu2024point2rbox}, especially in the following aspects: 
\textbf{1)} We rewrite the full text with a unified perspective and build a more complete and unified framework that enables support for multiple annotation formats among Point/HBox/RBox.
\textbf{2)} A more stringent theoretical foundation of symmetry-aware learning is elucidated to provide insight into why the network can discern object angles through consistency losses.
\textbf{3)} We technically simplify the paradigm with only one transformed view, resulting in a more concise architecture and a significant reduction in RAM usage.
\textbf{4)} The proposed Wholly-WOOD exhibits further improvements in accuracy compared to the conference versions, especially the performance of Point-to-RBox has increased by 22.36\%, benefiting from our new unified architecture and the newly devised P2R subnet. 
\textbf{5)} The model is applied to more Point/HBox-annotated scenarios, proving its effectiveness in reducing manual labeling in various applications.
\textbf{6)} We have released PyTorch and Jittor version codes for H2RBox-v2, Point2RBox, and Wholly-WOOD.}.
\textbf{2)} We propose symmetry-aware learning, a novel theory that leverages the reflection symmetry of visual objects to learn object angles through consistency losses, to address HBox-to-RBox conversion.
\textbf{3)} We propose the knowledge combination from synthetic visual patterns to handle Point-to-RBox conversion, utilizing synthetic patterns with known boxes to provide the necessary information for box regression.
\textbf{4)} Wholly-WOOD demonstrates superior accuracy compared to state-of-the-art methods in both HBox/Point-to-RBox settings.
\textbf{5)} We apply the model to various Point/HBox-annotated scenarios, showcasing its effectiveness in reducing manual labeling efforts in remote sensing and beyond.
\textbf{6)} The PyTorch \cite{Paszke2019PyTorch}\footnote{\url{https://github.com/yuyi1005/whollywood}.} and Jittor \cite{hu2020jittor}\footnote{\url{https://github.com/yuyi1005/whollywood-jittor}.} version codes for H2RBox-v2, Point2RBox, and Wholly-WOOD are released.

In a broader sense, the hope is that the dependence on costly manual annotation can be effectively mitigated, which could save a lot of human labor.

\section{Related work}\label{sec:work}

Beyond horizontal detection \cite{Zhao2019Object, liu2020deep,zheng2024zone}, oriented object detection (OOD) \cite{wen2023comprehensive} has received extensive attention. Here, approaches related to oriented detection and studies related to HBox/Point supervision are discussed.

\textbf{Fully-supervised oriented detection.} Representative works include anchor-based detector Rotated RetinaNet \cite{Lin2017Focal}, anchor-free detector Rotated FCOS \cite{Tian2019FCOS}, and two-stage solutions, e.g. RoI Transformer \cite{Ding2018Learning}, Oriented R-CNN \cite{Xie2021Oriented}, and ReDet \cite{Han2021Redet}. Some research enhances the detector by exploiting alignment features, e.g. R$^3$Det \cite{Yang2021R3Det} and S$^2$A-Net \cite{Han2022Align}. The angle regression may face boundary discontinuity and remedies are developed, including modulated losses \cite{Yang2019SCRDet, Qian2021RSDet} that alleviate loss jumps, angle coders \cite{Yang2020Arbitrary, Yang2021Dense, yu2024boundary} that convert the angle into boundary-free coded data, and Gaussian-based losses \cite{Yang2021Rethinking, Yang2021Learning, yang2023detecting, yang2023kfiou} transforming RBoxes into Gaussian distributions. RepPoint-based methods \cite{Yang2019Reppoints, hou2022grep, li2022oriented} provide alternatives that predict a set of sample points that bounds the spatial extent of an object. LMMRotate \cite{li2025simple} is a new paradigm of OOD based on multimodal language model and performs object localization through autoregressive prediction.

\textbf{HBox-to-RBox.} 
Before our studies, some methods use HBoxes with additional annotated data for training: \textbf{1)} OAOD \cite{iqbal2021leveraging} is proposed for weakly-supervised OOD. But in fact, it uses HBox along with an object angle as annotation, which is just ``slightly weaker" than RBox supervision. Such an annotation manner is not common, and OAOD is only verified on their self-collected ITU Firearm dataset. 
\textbf{2)} Sun et al. \cite{sun2021oriented} propose a two-stage framework: i) training detector with the annotated horizontal and vertical objects, and ii) mining the rotation objects by rotating the training image to align the oriented objects as horizontally or vertically as possible.
\textbf{3)} KCR \cite{zhu2023knowledge} combines a RBox-annotated source dataset with a HBox-annotated target dataset, and achieves HBox-to-RBox on the target dataset via transfer learning.

Some studies focus on a similar task, HBox-to-Mask: \textbf{1)} SDI \cite{khoreva2017simple} refines the segmentation through an iterative training process; \textbf{2)} BBTP \cite{hsu2019weakly} formulates the HBox-supervised instance segmentation into a multiple-instance learning problem based on Mask R-CNN \cite{he2017mask}; \textbf{3)} BoxInst \cite{tian2021boxinst} uses the color-pairwise affinity with box constraint under an efficient RoI-free CondInst \cite{tian2020conditional}; \textbf{4)} BoxLevelSet \cite{li2022box} introduces an energy function to predict the instance-aware mask as the level set; \textbf{5)} SAM (Segment Anything Model) \cite{kirillov2023segany} produces object masks from input Point/HBox prompts. Though RBoxes can be obtained from the segmentation mask by finding the minimum circumscribed rectangle, we show that such a cascade pipeline can be less cost-efficient (see Sec. \ref{sec:exp}). 

To fill the blank of HBox-to-RBox, we have proposed H2RBox \cite{yang2023h2rbox} and H2RBox-v2 \cite{yu2023h2rboxv2}. H2RBox directly achieves RBox detection from HBox annotations, bypassing segmentation. With HBox labels for the same object in various orientations, the geometric constraint limits candidate angles. Supplemented with a self-supervised branch eliminating the undesired results, an HBox-to-RBox paradigm is established. An enhanced version H2RBox-v2 \cite{yu2023h2rboxv2} is proposed to leverage the reflection symmetry of objects to estimate their angle, further boosting the HBox-to-RBox performance. Inspired by our work, EIE-Det \cite{wang2024explicit} uses an explicit equivariance branch for learning rotation consistency, and an implicit equivariance branch for learning position, aspect ratio, and scale consistency. AFWS \cite{lu2024afws} simplifies the model training process by decoupling horizontal and rotating parameters.

Particularly, our H2RBox-v2 \cite{yu2023h2rboxv2} has bridged the gap between HBox- and RBox-supervised OOD. In this paper, we employ a similar theoretical foundation in the HBox-to-RBox part of Wholly-WOOD, with a more concise architecture and significantly reduced RAM usage.

\textbf{Point-to-RBox.} Compared to Point-to-RBox, the Point-to-HBox setting has been better studied: \textbf{1)} P2BNet \cite{chen2022pointtobox} samples box proposals of different ratios and sizes around the labeled point and classifies them via multiple instance learning to achieve point-supervised horizontal object detection. \textbf{2)} PSOD \cite{gao2022weakly} achieves point-supervised salient object detection using an edge detector and adaptive masked flood fill. \textbf{3)} LESPS \cite{ying2023mapping} proposes a label evolution framework to progressively expand the point label by leveraging the intermediate predictions of CNNs for infrared small target detection.

Some methods accept partial point annotations (e.g. 80\% points and 20\% HBoxes), usually termed semi-supervision: 
\textbf{1)} Point DETR \cite{chen2021points} extends DETR \cite{carion2020detr} by adding a point encoder for point annotations. 
\textbf{2)} Group-RCNN \cite{zhang2022grouprcnn} generates a group of proposals for each point annotation.
\textbf{3)} CPR \cite{yu2022object} produces center points from coarse point annotations, relaxing the supervision from accurate points to freely spotted points.

Besides the Point-to-HBox methods, Point-to-Mask has also been an active area: Point2Mask \cite{li2023point2mask} is proposed to achieve panoptic segmentation using only a single point annotation per target for training. SAM (Segment Anything Model) \cite{kirillov2023segany} produces object masks from input Point/HBox prompts.

These Point-to-HBox/Mask methods are potentially applicable to our Point-to-RBox task setting -- by using a subsequent HBox/Mask-to-RBox to build a cascade solution. 

Recently, several approaches directly aimed at Point-to-RBox have been proposed: \textbf{1)} PointOBB \cite{luo2024pointobb} achieves point annotation based RBox generation method for oriented object detection through scale-sensitive consistency and multiple instance learning. \textbf{2)} P2RBox \cite{cao2023p2rbox} proposes oriented object detection with point prompts by employing the zero-shot Point-to-Mask ability of SAM \cite{kirillov2023segany}.

Our conference paper Point2RBox \cite{yu2024point2rbox} has also introduced a novel approach based on knowledge combination in this domain. While achieving competitive accuracy compared to state-of-the-art methods, it still has room for improvement, particularly in handling FPN/anchor assignments. In Wholly-WOOD, we incorporate the concept of knowledge combination and address the assignment issue, resulting in a substantially enhanced Point-to-RBox performance, about 22.36\%.

For comprehensive evaluation, our experiments will compare Wholly-WOOD with Point-to-RBox approaches such as PointOBB series \cite{luo2024pointobb,ren2025pointobbv2}, P2RBox \cite{cao2023p2rbox}, and Point2RBox \cite{yu2024point2rbox}, as well as cascade solutions driven by leading methodologies like P2BNet \cite{chen2022pointtobox} and Point2Mask \cite{li2023point2mask} (see Sec. \ref{sec:exp}). 

\section{Methods}\label{sec:met}

In this section, we delve into our series of research on weakly-supervised oriented detection. 
We begin in Sec. \ref{sec:met-sal} by presenting the foundational theory of symmetry-aware learning, demonstrating its ability to learn orientation from symmetry with theoretical guarantees. 
Next, Sec. \ref{sec:met-h2r} introduces H2RBox-v2, an implementation validating our theory and facilitating HBox-to-RBox conversion using symmetry-aware learning.
Leveraging the H2RBox-v2 pipeline, Sec. \ref{sec:met-p2r} illustrates Point2RBox, which employs synthetic pattern knowledge combination to achieve the Point-to-RBox conversion. 
Finally, we present Wholly-WOOD in Sec. \ref{sec:met-ww}, an integrated pipeline capable of accommodating diverse labeling formats (Points, HBoxes, RBoxes, and their combination), thereby offering an integral and adaptable solution.

\subsection{Theoretical guarantee of symmetry-aware learning}\label{sec:met-sal}

Assume there is a neural network $f_\text{nn}\left ( \cdot \right ) $ that maps a visual object $I$ to a real number $\theta$ representing the rotation:
\begin{equation}
\theta = f_\text{nn}\left ( I \right )
\end{equation}
where the visual object $I \in \mathbb{R}^{2\times M}$ is represented as a set of pixel locations; $M$ is the pixel count; $\theta \in \mathbb{R} \bmod \pi$, where $\theta_1 \equiv \theta_2 \pmod{\pi}$ implies $\theta_1 = \theta_2 + k\pi$ for some integer $k$.

In symmetry-aware learning, we simply train the network $f_\text{nn}\left ( \cdot \right ) $ to follow two properties, namely the flip consistency and the rotate consistency. 

\textbf{Property I: Flip consistency.} With an input object vertically flipped, $f_\text{nn}\left ( \cdot \right ) $ gives an opposite output:
\begin{equation}
- f_\text{nn}\left ( I \right ) \equiv f_\text{nn}\left ( \begin{bmatrix}
 1 & 0 \\
 0 & -1
\end{bmatrix} I \right ) \pmod{\pi}
\label{eq:flpcon} \\
\end{equation}

\textbf{Property II: Rotate consistency.} With an input rotated by $\mathcal{R}$, the output of $f_\text{nn}\left ( \cdot \right ) $ also rotates by $\mathcal{R}$:
\begin{equation}
f_\text{nn}\left ( I \right ) + \mathcal{R} \equiv f_\text{nn}\left ( \begin{bmatrix}
 \cos\mathcal{R} & -\sin\mathcal{R} \\
 \sin\mathcal{R} & \cos\mathcal{R}
\end{bmatrix} I \right ) \pmod{\pi}
\label{eq:rotcon}
\end{equation}

Here we provide a mathematical explanation for how the network can discern the angle of a reflective symmetric visual object through the rotate and flip consistencies. Let $\mathbf{x}$, $\mathbf{y}$ be perpendicular unit vectors in the plane. Suppose there exists a visual object, denoted as $I_\text{sym}$, which is reflection symmetric with a vector $\mathbf{u} = \cos\theta \mathbf{x}+\sin\theta \mathbf{y}$ representing the line of reflection. Based on the transformation matrix of reflection\footnote{\url{https://jonshiach.github.io/LA-book}.}, the reflection symmetry of $I_\text{sym}$ can be formulated as:
\begin{equation}
I_\text{sym} = \begin{bmatrix}
 \cos2\theta  & \sin2\theta \\
 \sin2\theta & -\cos2\theta
\end{bmatrix} I_\text{sym}
\label{eq:sym}
\end{equation}

By mapping the both sides of Eq. (\ref{eq:sym}) with the network function $f_\text{nn}\left ( \cdot \right )$, we obtain:
\begin{align}
f_\text{nn}\left ( I_\text{sym} \right ) &\equiv f_\text{nn}\left ( \begin{bmatrix}
 \cos2\theta  & \sin2\theta \\
 \sin2\theta & -\cos2\theta
\end{bmatrix} I_\text{sym} \right ) \\
&\equiv f_\text{nn}\left ( \begin{bmatrix}
 \cos2\theta  & -\sin2\theta \\
 \sin2\theta & \cos2\theta
\end{bmatrix}
\begin{bmatrix}
 1  & 0 \\
 0 & -1
\end{bmatrix} I_\text{sym} \right ) \label{eq:splitrf}\\
&\equiv -f_\text{nn}\left ( I_\text{sym} \right ) + 2\theta \pmod{\pi}
\label{eq:symfinal}
\end{align}
where Eq. (\ref{eq:splitrf}) indicates that a reflection transformation can be decomposed into the multiplication of a rotation and a flip. Substituting Eqs. (\ref{eq:flpcon}) and (\ref{eq:rotcon}) into Eq. (\ref{eq:splitrf}), we derive Eq. (\ref{eq:symfinal}). Solving Eq. (\ref{eq:symfinal}) yields:
\begin{equation}
f_\text{nn}\left ( I_\text{sym} \right ) \equiv \theta \pmod{\pi/2}
\label{eq:fnneqtheta}
\end{equation}
which suggests that IF: \textbf{1)} The input object $I_\text{sym}$ has reflection symmetry about the vector $\mathbf{u} = \cos\theta \mathbf{x}+\sin\theta \mathbf{y}$; AND \textbf{2)} $f_\text{nn}\left ( \cdot \right )$ subjects to the flip and rotate consistencies; THEN: $f_\text{nn}\left ( I_\text{sym} \right )$ precisely outputs the symmetry angle $\theta$ or an angle differing by $\pi/2$, which  is sufficient for learning rotation in OOD. 

Based on the above conclusion, training the network with flip and rotate consistencies leads to automatic regression of the object's angle in the network's output. Thereupon, we design a training pipeline to employ this approach in Sec. \ref{sec:met-h2r} and empirically confirm its effectiveness.

Notably, although the aforementioned study focuses on a single visual object, an assigner is employed to match objects in different views (detailed in Sec. \ref{sec:met-h2r}), enabling the calculation of consistency loss between these paired objects. Our theory can then be applied to each matched object center, extending the method to multiple object detection.

\subsection{H2RBox-v2}\label{sec:met-h2r}

\begin{figure*}[t]
\centering
\includegraphics[width=0.95\textwidth]{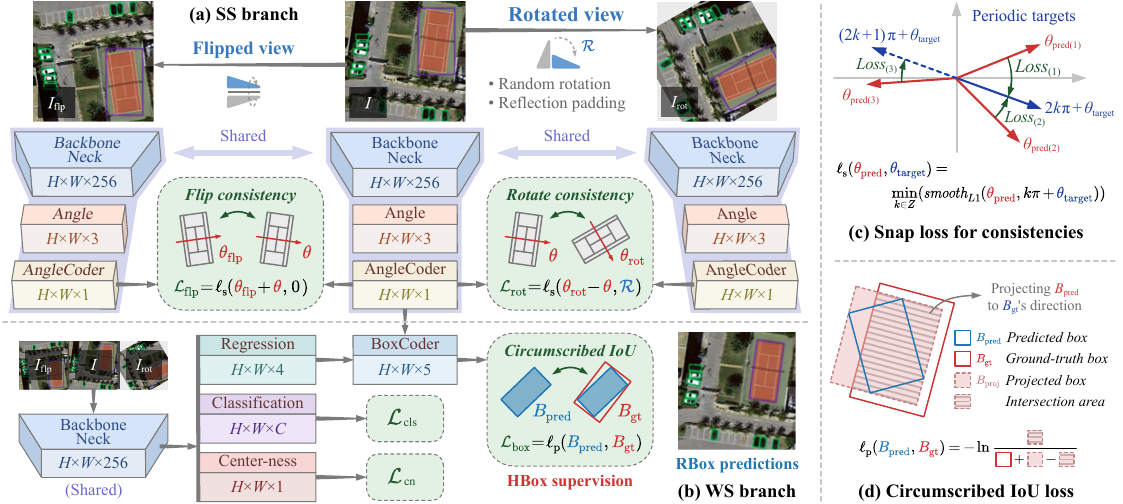}
\caption{The overview of H2RBox-v2. \textbf{(a)} Self-supervised (SS) branch that learns the orientation from the symmetry of objects. \textbf{(b)} Weakly-supervised branch that learns other properties from HBoxes. \textbf{(c)} Snap loss for the SS branch. \textbf{(d)} Circumscribed IoU (CircumIoU) loss for the WS branch.}\label{fig:h2r}
\end{figure*}

The training pipeline of the proposed H2RBox-v2 is given in Fig.~\ref{fig:h2r}, which consists of a self-supervised (SS) branch and a weakly-supervised (WS) branch. 

\textbf{Self-supervised (SS) branch.} It is designed to enforce the two consistencies by Eqs. (\ref{eq:flpcon}) and (\ref{eq:rotcon}) within the neural network. As shown in Fig.~\ref{fig:h2r}a, we perform vertical flip and random rotation to generate two transformed views, $I_\text{flp}$ and $I_\text{rot}$, of the input image $I$. The blank border area induced by rotation is filled with reflection padding. Then the three views are fed into three parameter-shared branches of the network, where ResNet50 \cite{He2016Deep} and FPN \cite{Lin2017Feature} are used as the backbone and the neck, respectively. The random rotation is in the range $\pi/4\sim3\pi/4$ (according to the ablation in Table \ref{tab:abl-range}).

Next, a label assigner is required to match the objects in different views. We use the default center sampling assigner of FCOS detector to calculate the average angle features on all sample points for each object and eliminate those objects without correspondence (lost during rotation). 

Following the assigner, PSC \cite{yu2023psc} angle coder is adopted to cope with the boundary problem. We empirically demonstrate in Table \ref{tab:abl-ssloss} that PSC is necessary to achieve a stable convergence of training. The output angles of the original, flipped, and rotated views are denoted as $\theta$, $\theta_\text{flp}$, and $\theta_\text{rot}$.

Then, the losses for the consistencies can be expressed as:
\begin{equation}
\left\{\begin{array}{l}
\mathcal{L}_\text{flp} = \ell_\text{s}  \left ( \theta _\text{flp} + \theta, 0 \right )  \\
\mathcal{L}_\text{rot} = \ell_\text{s}  \left (  \theta _\text{rot} - \theta, \mathcal{R}  \right )
\end{array}\right.
\label{eq:loss_flp_rot}
\end{equation}
where $\mathcal{L}_\text{flp}$ is the loss for flip consistency and $\mathcal{L}_\text{rot}$ for rotate consistency. $\mathcal{R}$ is the rotation angle in the rotated view generation. During the calculation of Eq. (\ref{eq:loss_flp_rot}), $\ell_\text{s}\left ( \cdot \right )$ named snap loss\footnote{There are a series of targets with interval $\pi$, just like a series of evenly spaced grids. The snap loss moves prediction toward the closest target, thus deriving its name from the ``snap to grid" function.} (see Fig.~\ref{fig:h2r}c) is proposed as:
\begin{equation}
\ell_s\left ( \theta_\text{pred}, \theta_\text{target} \right ) =\min_{k\in Z} \left ( smooth_{L1}\left ( \theta_\text{pred}, k\pi+\theta_\text{target}  \right ) \right )
\end{equation}
where the $\min\left ( \cdot \right )$ operation regresses the prediction toward the closest target to circumvent the periodicity problem (see Table \ref{tab:abl-ssloss} for ablation). This equation can be done by using the modulo operation in the code implementation.

Finally, the loss for the SS branch can be expressed as:
\begin{equation}
\mathcal{L}_\text{ss} = \mathcal{L}_\text{rot} + \lambda \mathcal{L}_\text{flp}
\end{equation}
where $\lambda$ adjusts the weight between rotation and flip, set to 0.05 according to the ablation in Table \ref{tab:abl-ssloss}.

By minimizing $\mathcal{L}_\text{ss}$, the network learns to conform with flip and rotate consistencies and gains the ability of angle prediction through self-supervision.

\textbf{Weakly-supervised (WS) branch.} To predict other properties of the bounding box (position, size, category, etc.), a weakly-supervised branch using HBox supervision is supplemented, as shown in Fig.~\ref{fig:h2r}b. The losses to learn these properties are mainly defined by the backbone FCOS detector, including $\mathcal{L}_\text{cls}$ for classification and $\mathcal{L}_\text{cn}$ for center-ness. 

In our WS task setting, the ground-truth box $B_\text{gt}$ is an HBox/RBox circumscribed to the predicted box $B_\text{pred}$. Therefore, we design a circumscribed IoU (CircumIoU) for the box regression (see Fig. \ref{fig:h2r}d) as: 
\begin{equation}
\mathcal{L}_\text{box} = \ell_p\left ( B_\text{pred}, B_\text{gt} \right ) = -\ln \frac{B_\text{proj} \cap B_\text{gt}}{B_\text{proj} \cup B_\text{gt}}
\end{equation}
where $B_\text{proj}$ is the dashed box in Fig.~\ref{fig:h2r}d, obtained by projecting the predicted box $B_\text{pred}$ to the direction of $B_\text{gt}$. 

\textbf{Overall loss.} The overall loss for H2RBox-v2 is:
\begin{equation}
\mathcal{L}_\text{h2rbox-v2} = \mathcal{L}_\text{cls} + \mu_\text{cn} \mathcal{L}_\text{cn} + \mu_\text{box} \mathcal{L}_\text{box} + \mu_\text{ss} \mathcal{L}_\text{ss}
\end{equation}
where $\mu_\text{cn}$, $\mu_\text{box}$, and $\mu_\text{ss}$ are set to one by default.

\subsection{Point2RBox}\label{sec:met-p2r}

\begin{figure*}[t]
\centering
\includegraphics[width=0.9\textwidth]{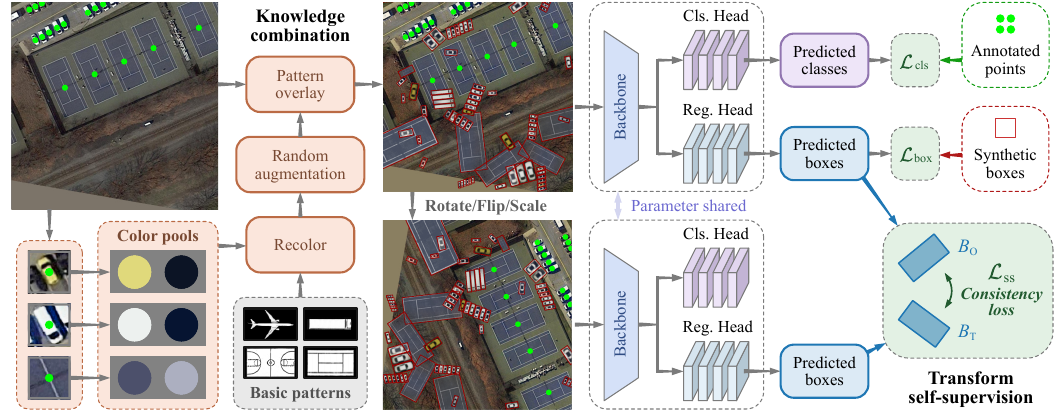}
\caption{The training flowchart of Point2RBox, consisting of knowledge combination and transform self-supervision. The core idea is to combine knowledge from synthetic patterns for size and angle estimation, and knowledge from annotated points for classification.}\label{fig:p2r}
\end{figure*}

Based on our HBox-to-RBox pipeline, we further devise the flowchart in Fig. \ref{fig:p2r} for point-supervised rotated detection. 

\textbf{Knowledge combination.} During manual annotation, annotators are often provided with a one-shot example for each category. For point annotations, the exact size and angle of the labeled object are unknown, but the example allows us to generate similar patterns. Since these patterns are derived from a known example, their bounding boxes are also known (see red RBoxes in Fig. \ref{fig:p2r}), providing the necessary information for box regression. Building upon this concept, the knowledge combination module is devised. First, we sample around each labeled point, and extract its neighbor colors, namely the face color $C_\text{face}$ and the edge color $C_\text{edge}$, as follows:
\begin{equation}
\left\{\begin{array}l
 C_\text{face}=\text{mean}\left ( I_0  \right ) 
 \\
 C_\text{edge}=\text{sum}\left ( dI_1 \right ) 
\end{array}\right.
\end{equation}
where $I_0$ and $I_1$ are the neighbor pixels around a labeled point. We simply use a $5\times5$ neighbor area for $I_0$ and $33\times33$ for $I_1$. Here $d$ is the gradient of $I_1$ indicating the edge intensity of each pixel (the sum of $d$ is uniform to one).

Then, we spread the two extracted colors to a basic pattern. The basic pattern is a gray-scale sample manually cropped from training images and adjusted to gray-scale (one sample for each category), which can be denoted as $P$, with its value in the range $(0, 1)$. The recolor step can be expressed as:
\begin{equation}
P_\text{recolor}=PC_\text{face}+\left ( 1-P \right ) C_\text{edge}
\end{equation}

Such an ``extract-and-spread" design has two advantages: \textbf{1)} The diversity of the synthetic patterns is significantly enriched. \textbf{2)} The gap between generated patterns and real ones is narrowed. By this means, the knowledge can be better transferred to estimate the RBoxes of the real objects (see ablation in Sec. \ref{sec:exp-abl}). 

Afterward, the recolored patterns are augmented with the random flip, resize, and rotation, and moved to a random position inside the image border. The probability for random flip and rotation is set to 0.5 and 1, respectively. The random resize can be formulated as:
\begin{equation}
\setlength{\abovedisplayskip}{6pt}
\setlength{\belowdisplayskip}{6pt}
\left\{\begin{array}l
w = w_0 \exp\left ( \sigma _\text{base} + \sigma_w  \right )  \\
h = h_0 \exp\left ( \sigma _\text{base} + \sigma_w + \sigma_r \right )
\end{array}\right.
\end{equation}
where $\sigma _\text{base}$ is a random number from the standard normal distribution $\mathcal{N}(0, 0.4)$ for each image; $\sigma_w$ and $\sigma_r$ are random numbers drawn independently from the same distribution for each instance; $w_0$ and $h_0$ are the original pattern sizes; $w$ and $h$ are the resized ones.

To avoid overlapping patterns, NMS (Non-Maximum Suppression) is then applied so that the IoU between synthetic patterns is less than 0.05. Furthermore, to avoid the real objects being completely occluded, transparent blending is used:
\begin{equation}
\alpha(x,y)=\alpha_1\exp\left ( -k_0 x^2-k_1 y^2 \right )  +\alpha_0
\end{equation}
where $\alpha(x,y)$ is the opacity channel of the synthetic pattern; $x$ and $y$ are coordinates in range $[-1, 1]$; $k_0$ and $k_1$ are random numbers in $[0.1, 2]$ from uniform distribution; $\alpha_0=0.1$ and $\alpha_1=0.9$ keep the opacity between 10\% to 100\%.

Finally, these generated patterns are overlaid on the original image and their known bounding boxes are used for training, providing the knowledge for box regression.

\textbf{Label assignment.} Available detectors largely rely on FPN (Feature Pyramid Network) \cite{Lin2017Feature} or anchors of various scales to deal with objects of different sizes. For example, \textbf{1)} Rotated FCOS \cite{Tian2019FCOS} uses five feature layers, with large and small objects assigned to different ones. \textbf{2)} YOLOF \cite{chen2021yolof} presents the one-level feature layer, but it still uses five preset anchors with sizes 32, 64, 128, 256, and 512.

While point annotations do not provide any size information, they do not apply to such an FPN/anchor-based assignment strategy. Therefore, we use YOLOF as the backbone detector with all five anchors set to a fixed size ($64 \times 64$ for the DOTA dataset and $128 \times 128$ for the others). 

Instead of assigning ground-truths to the anchor with the highest IoU, we assign them (including both labeled points and synthetic boxes) to the one that produces the highest classification score. Then the matching scores between anchors and ground-truths can be calculated as:
\begin{equation}
\setlength{\abovedisplayskip}{6pt}
\setlength{\belowdisplayskip}{6pt}
\mathit{ score}= \begin{cases}
0, \quad\quad L_1\left ( xy_\text{pred},xy_\text{gt} \right )>32\\
c_\text{pred} , \quad\mathit{otherwise}
\end{cases} 
\end{equation}
where $xy_\text{pred}$ and $xy_\text{gt}$ are the center coordinates of predicted boxes and ground-truths; $c_\text{pred}$ is the predicted classification scores corresponding to the labels. Afterward, following the setting of YOLOF, we use K-nearest to find four positive anchors with the highest scores for each ground truth. 

\textbf{Transform self-supervision.} 
Drawing from the effective approach validated in H2RBox-v2, we also perform self-supervision within Point2RBox. In addition to the rotation and flip views, we now incorporate a scale view as well. To reduce RAM usage, instead of utilizing three concurrent views, we employ a single view randomly chosen from a distribution of transformations: 66.5\% rotation, 3.5\% flipping, and 30\% scaling (partly based on $\lambda = 0.05$ in Sec. \ref{sec:met-h2r}).

When the input image is scaled by $s$, the center coordinates and the size of output RBoxes should be likewise scaled. Thus the self-supervised loss for the scale view is:
\begin{equation}
\setlength{\abovedisplayskip}{6pt}
\setlength{\belowdisplayskip}{6pt}
\mathcal{L}_\text{sca}  = \mathit{GIoU}  \left ( \mathit{r2h} (B _\text{ori}) \times s, \mathit{r2h}(B_\text{trs}) \right ) 
\label{eq:loss_sca}
\end{equation}
where $B _\text{ori}$ and $B _\text{trs}$ are outputs of the original and scale views; $\mathit{r2h}\left(\cdot\right)$ is the function to get circumscribed HBoxes, $s$ is the scaling factor applied to the input image in range $\left( 0.5, 1.5 \right)$.

The loss of self-supervision can be expressed as:
\begin{equation}
\mathcal{L}_\text{ss} = \mathcal{L}_\text{rot} + \mu_\text{flp}\mathcal{L}_\text{flp} + \mu_\text{sca}\mathcal{L}_\text{sca}
\end{equation}
where $\mu_\text{flp}$ and $\mu_\text{sca}$ are set to one by default in this paper.

\textbf{Overall loss.} Point annotations are only used to train the classification, and the loss $\mathcal{L}_\text{cls}$ to learn the classification is defined by the backbone YOLOF detector. Known boxes of synthetic patterns are used to train the box regression, and the loss is calculated with RotatedIoU \cite{zhou2019iou, zheng2020rotation}:
\begin{equation}
\mathcal{L}_\text{box} = -\ln \frac{M_\text{box}B_\text{pred} \cap M_\text{box}B_\text{gt}}{M_\text{box}B_\text{pred} \cup M_\text{box}B_\text{gt}}
\label{eq:p2rlossbox}
\end{equation}
where $M_\text{box}$ is a mask to select RBoxes that are assigned to synthetic patterns.

The overall loss for Point2RBox can be expressed as:
\begin{equation}
\mathcal{L}_\text{point2rbox} = \mathcal{L}_\text{cls} + \mu_\text{box} \mathcal{L}_\text{box} + \mu_\text{ss} \mathcal{L}_\text{ss}
\end{equation}
where $\mu_\text{box}$ and $\mu_\text{ss}$ are set to one by default.

\subsection{Wholly-WOOD}\label{sec:met-ww}

\begin{figure*}[t]
\centering
\includegraphics[width=0.9\textwidth]{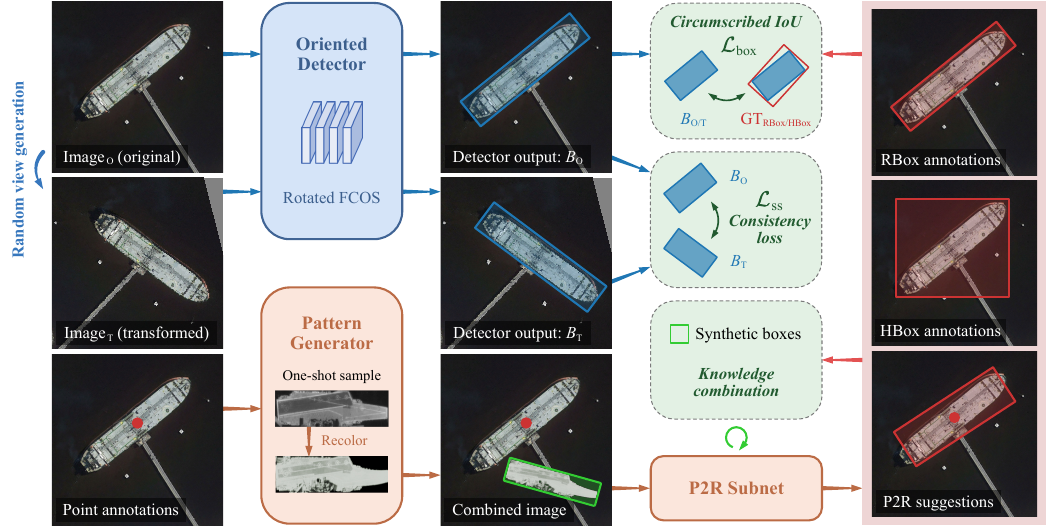}
\caption{The Wholly-WOOD architecture. It features two key components: \textbf{1)} The symmetry-aware learning module utilizes self-supervised learning to extract object orientations based on symmetry; \textbf{2)} The knowledge combination module integrates a pattern generator, which generates synthetic visual patterns for training the size and angle regression of the P2R subnet. The predictions from the P2R subnet then offer RBox suggestions corresponding to each point.}\label{fig:wholly}
\end{figure*}

Based on the principles in H2RBox-v2 and Point2RBox, an integral and comprehensive solution, Wholly-WOOD, is proposed to wholly leverage diversified-quality labels including RBoxes, HBoxes, Points, or their combination. 

The schematic representation of Wholly-WOOD is presented in Fig. \ref{fig:wholly}. The upper part (blue arrows) is derived from H2RBox-v2, which plays a crucial role in HBox-to-RBox. The lower part is derived from Point2RBox to process point annotations, where a pattern generator (the same as that in Point2RBox) is used to generate synthetic visual patterns. By training the P2R subnet, it combines the knowledge from these patterns to generate RBox suggestions of point-annotated objects. Afterward, these P2R suggestions, together with RBox/HBox annotations, are harnessed by the upper part to train the Rotate FCOS detector.

To eliminate the limitations (i.e. the FPN/anchor assignment issue) of Point2RBox, we propose the P2R subnet to replace the YOLOF detector. Below, we introduce the SS and WS branches of Wholly-WOOD, followed by a detailed description of the newly devised P2R subnet.

\textbf{SS branch and WS branch.} Similar to the H2RBox-v2, Wholly-WOOD also consists of the SS branch and the WS branch. The SS branch is designed to enforce the two consistencies for symmetry-aware learning. We employ a single view randomly chosen from a distribution of transformations: 95\% rotation and 5\% flipping (based on $\lambda = 0.05$ in Sec. \ref{sec:met-h2r}). When rotation is applied, the loss $\mathcal{L}_\text{rot}$ is computed to measure the disparity between the outputs of the two views. Similarly, $\mathcal{L}_\text{flp}$ is utilized to assess flip-induced variations. When the network adheres to the two consistencies, it automatically gains the ability to predict the angle of objects (Refer to Sec. \ref{sec:met-sal} for the explanation).

The loss for the SS branch can be expressed as:
\begin{equation}
\mathcal{L}_\text{ss} = \mathcal{L}_\text{rot} + \mu_\text{flp} \mathcal{L}_\text{flp}
\end{equation}
where $\mu_\text{flp}=1$ by default as the weight between rotation and flip has been featured by the proportion of view generation.

Meanwhile, the WS branch is used to process HBox/RBox annotations. CircumIoU (Fig. \ref{fig:h2r}d) is used for HBoxes circumscribed to the predicted boxes and RotatedIoU for RBoxes. The overall loss for Wholly-WOOD can be expressed as:
\begin{equation}
\mathcal{L}_\text{wholly-wood} = \mathcal{L}_\text{cls} + \mu_\text{cn} \mathcal{L}_\text{cn} + \mu_\text{box} \mathcal{L}_\text{box} + \mu_\text{ss} \mathcal{L}_\text{ss}
\end{equation}
where $\mu_\text{cn}$, $\mu_\text{box}$, and $\mu_\text{ss}$ are set to one by default.

\textbf{P2R subnet.} As mentioned in Sec. \ref{sec:met-p2r}, objects annotated with points cannot be assigned to different FPN layers or anchors based on their sizes. However, available detectors including FCOS \cite{Tian2019FCOS} and YOLOF \cite{chen2021yolof} rely on FPN layers or multiple anchors to deal with objects of different sizes. In Point2RBox, we simply use YOLOF with a fixed anchor size, which partly circumvents the issue but also limits the box regression range, leading to insufficient accuracy. 

\begin{figure*}[t]
\centering
\includegraphics[width=0.95\textwidth]{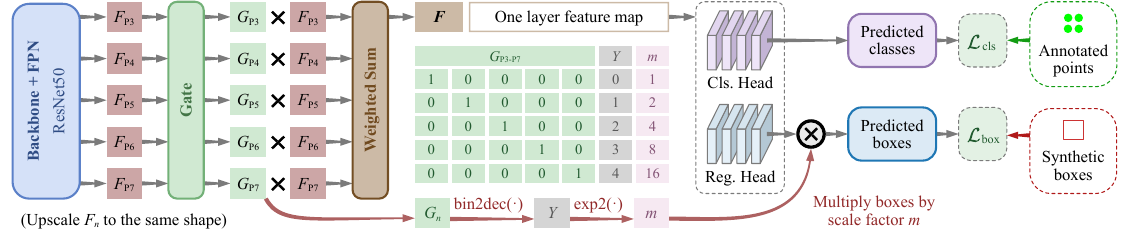}
\caption{The architecture of P2R subnet. The multiple output layers of the FPN automatically aggregate based on the gating score activated by each layer itself. Meanwhile, the output boxes are scaled by a factor calculated from the gating score. bin2dec($\cdot$) denotes continuous binary-to-decimal (see Eq. \ref{eq:b2d}).}\label{fig:p2rs}
\end{figure*}

To further address this problem, we devise a novel ``fusion and scaling" mechanism (see Fig. \ref{fig:p2rs}). The P2R subnet is based on FCOS with ResNet50 \cite{He2016Deep} and FPN \cite{Lin2017Feature}. It is anchor-free with only one feature layer, yet it allows the prediction of both large and small objects. Specifically, the multiple output layers of the FPN are automatically aggregated based on a self-activated gating score:
\begin{equation}
G_n = \mathit{softmax}\left ( \mathit{conv} \left ( \mathit{interp} \left ( F_n \right ) \right ) \right )
\label{eq:gn_fn}
\end{equation}
where $F_n$ is the $n$-th FPN feature layer; $\mathit{interp}\left ( \cdot \right )$ upscales $F_n$ to the shape of $F_1$ though nearest interpolation; $G_n$ is the gating score for each layer; $\mathit{conv}\left ( \cdot \right )$ is a $3 \times 3$ convolution layer with one output channel; $\mathit{softmax}\left ( \cdot \right )$ normalizes the sum of $G_1, G_2, \cdots, G_N$ to one at each pixel.

The final one-layer feature map $F$ can be obtained by:
\begin{equation}
F =  {\textstyle \sum_{n=1}^{N}} G_n \cdot \mathit{interp} \left ( F_n \right ) 
\label{eq:f_gn}
\end{equation}
where $N$ is the total number of FPN layers, $N=5$ (P3 to P7 layers) by default. Equations (\ref{eq:gn_fn}) and (\ref{eq:f_gn}) indicate that the weights to fuse the layers are generated by each layer itself.

Meanwhile, $G_n$ is treated as an $N$ bit binary to calculate a scale factor (see Fig. \ref{fig:p2r}). We first calculate $Y$ as:
\begin{equation}
Y=\frac{N}{2\pi}\left(\pi-\arctan \frac{\sum_{n=1}^{N} G_{n} \sin \left(2\pi\cdot\frac{n-1}{N}\right)}{\sum_{n=1}^{N} -G_{n} \cos \left(2\pi\cdot\frac{n-1}{N}\right)} \right)
\label{eq:b2d}
\end{equation}
where the calculation is essentially a binary decoder \cite{yu2024boundary} that decodes an $N$ bit binary to decimal in a continuous manner. 

The scale factor in range $\left [1, 2^N \right)$ is then calculated by $m=2^Y$ and the size of bounding box is predicted as:
\begin{equation}
B_\text{pred} = m \cdot \mathit{conv}\left ( F \right ) 
\label{eq:boxtimeslambda}
\end{equation}
where $B_\text{pred}$ is the prediction for the size of the bounding box; $\mathit{conv}\left ( \cdot \right ) $ is the convolution layer with four output channels for size regression. Our ablation study (see Fig. \ref{fig:gate}) demonstrates that different objects have varying scale factors, which significantly extend the dynamic range of box regression.

During the label assignment, we simply assign the point 
annotations or the center of synthetic patterns to the nearest pixel on the one-layer feature map $F$. 

\textbf{Advantages of the new design.} Such a design can automatically deal with objects of different sizes. Most importantly, it is anchor-free and has only one feature layer, circumventing the assignment issue for point annotations. Benefiting from this novel mechanism and the accordingly devised P2R subnet, Wholly-WOOD achieves a much higher AP$_{50}$ accuracy in Point-to-RBox conversion compared to our conference work Point2RBox \cite{yu2024point2rbox} (62.63\% vs. 40.27\%, see Table \ref{tab:dotav1}).

\textbf{Loss for P2R subnet.} The subnet is trained parallel to the main detector. The overall loss for the P2R subnet is as:
\begin{equation}
\mathcal{L}_\text{p2r-subnet} = \mathcal{L}_\text{cls} + \mu_\text{box}\mathcal{L}_\text{box}
\end{equation}
where $\mu_\text{box}$ is set to one by default; $\mathcal{L}_\text{cls}$ is the classification loss defined by the FCOS detector; $\mathcal{L}_\text{box}$ is defined by Eq. (\ref{eq:p2rlossbox}).

\subsection{Inference procedure}

In all our devised pipelines (i.e. H2RBox, H2RBox-v2, Point2RBox, and Wholly-WOOD), the self-supervision is solely utilized during training. During inference, there is no requirement for self-supervision or view generation, and only the forward propagation of the detector is involved. As a result, these methods have similar inference speeds compared to the backbone detector on which they are based.

\section{Experiments}\label{sec:exp}

\subsection{Datasets, settings, and metrics}

\textbf{Datasets.} To evaluate our approach, we assess its performance using five remote sensing datasets: DOTA, HRSC, FAIR1M, SARDet-100K, and STAR. These datasets are originally annotated with RBoxes, from which we derive Points or HBoxes by extracting the center point or minimum circumscribed rectangle respectively. These Points/HBoxes serve as input for Wholly-WOOD, and the resulting outputs are compared against RBox-trained counterparts to evaluate performance disparities (see Tables \ref{tab:dotav1} and \ref{tab:otherds}). Afterward, we apply Wholly-WOOD to datasets annotated only with Points/HBoxes to showcase the practical effectiveness of our method. The experiments span multiple scenarios, including Synthetic Aperture Radar (SAR) images, microscope images, and Printed Circuit Board (PCB) images (see Fig. \ref{fig:expa}).

\textbf{1)} DOTA \cite{Xia2018DOTA}: DOTA-v1.0 contains 2,806 aerial images, 1,411 for training, 937 for validation, and 458 for testing, as annotated using 15 categories with 188,282 instances in total. DOTA-v1.5/2.0 are the extended version of v1.0.
We follow the default preprocessing in MMRotate \cite{Zhou2022MMRotate}: The high-resolution images are split into 1,024 $\times$ 1,024 patches with an overlap of 200 pixels for training, and the detection results of all patches are merged to evaluate the performance.

\textbf{2)} HRSC \cite{Liu2017HRSC}: It contains ship instances both on the sea and inshore, with arbitrary orientations. The training, validation, and testing set includes 436,181, and 444 images, respectively. With preprocessing by MMRotate, images are scaled to 800 $\times$ 800 for training/testing.

\textbf{3)} FAIR1M \cite{Sun2022FAIR1M}: It contains more than one million instances for fine-grained object recognition in high-resolution remote sensing imagery. The dataset is annotated with 37 fine-grained categories. We split the images into 1,024 $\times$ 1,024 patches with an overlap of 200 pixels and a scale rate of 1.5 and merge the results for testing on the FAIR1M-1.0 server.

\textbf{4)} SARDet-100K \cite{li2024sardet100k}: It is a large-scale Synthetic Aperture Radar (SAR) object detection dataset, containing six categories and more than 100 thousand instances. The dataset provides HBox annotations only, and we use it to verify if our model can build an RBox version from it.

\textbf{5)} STAR \cite{li2024star}: The dataset is extensive for scene graph generation, covering more than 210,000 objects with diverse spatial resolutions, classified into 48 fine-grained categories and precisely annotated with oriented bounding boxes. 

\textbf{Settings.} Using PyTorch 1.13.1 \cite{Paszke2019PyTorch} and the rotation detection tool kits: MMRotate 1.0.0 \cite{Zhou2022MMRotate}, experiments are carried out. The performance comparisons are obtained by using the same platforms (i.e. PyTorch/MMRotate version) and hyper-parameters (learning rate, batch size, optimizer, etc.).

We adopt the FCOS \cite{Tian2019FCOS} detector with ResNet50 \cite{He2016Deep} backbone and FPN \cite{Lin2017Feature} neck as the baseline, based on which we develop our unified detector. All models are trained with AdamW \cite{loshchilov2018decoupled}, with an initial learning rate of 5e-5 and a mini-batch size of 2, on NVIDIA RTX3090/4090 GPUs. We adopt a learning rate warm-up for 500 iterations, and the rate is divided by ten at each decay step. ``1$\times$" and ``6$\times$" schedules indicate 12 and 72 epochs for training. ``MS" and ``RR" denote multi-scale technique \cite{Zhou2022MMRotate} and random rotation augmentation. Unless otherwise specified, ``6$\times$" is used for HRSC and ``1$\times$" for the other datasets, while random flipping is the only augmentation as always adopted by default.

\textbf{Metrics.} We choose Average Precision (AP), a commonly used metric in object detection tasks, as the primary metric. It quantifies the accuracy of a model in identifying objects within an image. AP is calculated by measuring the area under the precision-recall curve, where ``precision" is the ratio of true positives to the sum of true positives and false positives. The detected box is considered correct when the Intersection over Union (IoU) between the detected box and the ground truth is no less than 50\% (denoted in subscript as AP$_{50}$). The metric ranges from 0 to 1, with higher values indicating better performance. In multi-class detection, the AP is averaged across different classes to obtain the mean average precision, providing an overall performance measure for the model. 

\subsection{Ablation studies}\label{sec:exp-abl}

\textbf{Boundary problem.} Table \ref{tab:abl-ssloss} studies the impact of using the snap loss (see Sec.~\ref{sec:met-h2r}) and the angle coder on H2RBox-v2. Column ``PSC" indicates using PSC angle coder \cite{yu2023psc} and ``w/o PSC" means the conv layer directly outputs the angle. Column ``$\ell_\text{s}$" with check mark denotes using snap loss (otherwise using smooth L1 loss). Without these two modules handling boundary discontinuity, we empirically find that the loss could fluctuate in a wide range, even failure in convergence (see the much lower results in Table \ref{tab:abl-ssloss}). In comparison, when both PSC and snap loss are used, the training is stable.

\textbf{CircumIoU for WS branch.} Table \ref{tab:abl-wsloss} shows that CircumIoU loss is compatible with random rotation augmentation (RR) to further improve the performance, which H2RBox is incapable of.  ``$\ell_\text{p}$" means using CircumIoU loss in Sec.~\ref{sec:met-h2r}, and otherwise, IoU loss \cite{yu2016unitbox} is used following a conversion from RBox to HBox (refer to H2RBox \cite{yang2023h2rbox}). 

\textbf{Weights between }$\mathcal{L}_\text{flp}$\textbf{ and }$\mathcal{L}_\text{rot}$\textbf{.} Table \ref{tab:abl-weight} shows that on both DOTA and HRSC datasets, $\lambda = 0.05$ could be the best choice under AP$_{50}$ metric, whereas $\lambda = 0.1$ under AP$_{75}$. Hence in most experiments, we choose $\lambda = 0.05$, except for Table \ref{tab:abl-range} where $\lambda = 0.1$ is used. Following $\lambda = 0.05$ in H2RBox-v2, we employ 95\% rotation and 5\% flipping in the random view generation of Wholly-WOOD.

\begin{table}[!tb]
\caption{Ablation of using PSC coder and snap loss to address the stability issue in symmetry-aware learning.}
\label{tab:abl-ssloss}
\setlength{\tabcolsep}{4mm}
\centering
\begin{tabular}{c|cc|ccc}
\hline
\rule{0pt}{9pt}\textbf{Dataset}      & PSC        & $\ell_\text{s}$ & \textbf{AP}    & \textbf{AP$_{50}$}  & \textbf{AP$_{75}$}  \\
\hline
\multirow{4}{*}{DOTA} &            &            & 24.24 & 52.24 & 19.48 \\
                      &            & \checkmark & 0.01  & 0.77  & 0.02  \\
                      & \checkmark &            & 10.49 & 27.57 & 6.15  \\
                      & \checkmark & \checkmark & \textbf{40.69} & \textbf{72.31} & \textbf{39.49} \\
\hline
\multirow{4}{*}{HRSC} &            &            & 2.25  & 7.83  & 0.62  \\
                      &            & \checkmark & 48.95 & 88.52 & 50.03 \\
                      & \checkmark &            & 0.31  & 0.88  & 0.13  \\
                      & \checkmark & \checkmark & \textbf{58.03} & \textbf{89.66} & \textbf{64.80}  \\
\hline
\end{tabular}
\end{table}

\begin{table}[!tb]
\caption{Ablation of using CircumIoU loss and random rotation augmentation (RR) in the WS branch.}
\label{tab:abl-wsloss}
\setlength{\tabcolsep}{4.05mm}
\centering
\begin{tabular}{c|cc|ccc}
\hline
\rule{0pt}{9pt}\textbf{Dataset}               & \,$\ell_\text{p}$   & RR        & \textbf{AP}    & \textbf{AP$_{50}$}  & \textbf{AP$_{75}$}  \\
\hline
\multirow{4}{*}{DOTA} &   &    & 39.35 & 71.49 & 37.03 \\
                      &   &  \checkmark  & 11.93 & 29.34 & 7.86 \\
                      & \checkmark  &    & \textbf{40.69} & \textbf{72.31} & 39.49 \\
                      & \checkmark  &  \checkmark  & 40.17 & 71.79 & \textbf{39.77} \\
\hline
\multirow{4}{*}{HRSC} &   &    & 56.20  & 89.58 & 61.84 \\
                      &   &  \checkmark  & 41.10     & 87.19     & 33.97     \\
                      &  \checkmark &    & 58.03 & \textbf{89.66} & 64.80  \\
                      &  \checkmark &  \checkmark  & \textbf{63.82} & 89.56 & \textbf{76.11}  \\
\hline
\end{tabular}
\end{table}

\begin{table}[!tb]
\caption{Ablation with different weights between flipping and rotating losses defined in Eq.~\ref{eq:loss_flp_rot}. }
\label{tab:abl-weight}
\setlength{\tabcolsep}{5.0mm}
\centering
\begin{tabular}{c|c|ccc}
\hline
\rule{0pt}{9pt}\textbf{Dataset} & $\lambda$ & \textbf{AP}    & \textbf{AP$_{50}$}  & \textbf{AP$_{75}$} \\
\hline
\multirow{6}{*}{DOTA}    & 0   & 31.60 & 66.37 & 25.03 \\
& 0.01   & 40.43 & 72.26 & 38.55 \\
& 0.05   & \textbf{40.69} & \textbf{72.31} & 39.49 \\
& 0.1    & 40.48 & 71.46 & \textbf{39.84} \\
& 0.5    & 39.94 & 72.26 & 38.16 \\
& 1.0    & 38.50  & 70.91 & 36.02 \\
\hline
\multirow{6}{*}{HRSC}    & 0   & 0.06 & 0.32  & 0.00 \\
& 0.01   & 55.78 & 89.20  & 61.72 \\
& 0.05   & 58.03 & \textbf{89.66} & 64.80 \\
& 0.1    & \textbf{58.22} & 89.45 & \textbf{64.99} \\
& 0.5    & 53.85 & 88.90  & 61.47 \\
& 1.0    & 1.57  & 6.97  & 0.38  \\
\hline
\end{tabular}
\end{table}

\textbf{Range of view generation.} When the rotation angle $\mathcal{R}$ is close to 0, the SS branch could fall into a sick state. This may explain the fluctuation of losses under the random rotation within $-\pi\sim\pi$, leading to training instability. According to Table \ref{tab:abl-range}, $\pi/4\sim3\pi/4$ is more suitable.

\textbf{View multiplexing.} Wholly-WOOD employs a single view randomly chosen from 5\% flip or 95\% rotation (the proportion based on $\lambda=0.05$ in Table \ref{tab:abl-weight}). Compared to H2RBox-v2, such a multiplexing design of Wholly-WOOD achieves higher AP$_{50}$ (73.48\% vs. 72.31\%) while significantly reducing the training time and the RAM usage.

\begin{table}[!tb]
\caption{Ablation with different random ranges in the rotated view generation on HRSC. }
\label{tab:abl-range}
\setlength{\tabcolsep}{5.4mm}
\centering
\begin{tabular}{c|ccc}
\hline
\rule{0pt}{9pt}\textbf{Range}   & \textbf{AP} & \textbf{AP$_{50}$} & \textbf{AP$_{75}$}  \\
\hline
$-\pi\sim\pi ^*$      & 56.57 & 89.47   & 63.14   \\
$\pi/4\sim3\pi/4$     & \textbf{58.22} & 89.45   & \textbf{64.99}   \\
$3\pi/8\sim5\pi/8$    & 56.81 & \textbf{89.83}   & 64.03   \\
$7\pi/16\sim9\pi/16$  & 55.56 & 89.40   & 61.28  \\
\hline
\specialrule{0pt}{2pt}{0pt}
\multicolumn{4}{l}{$^*$Not stable, occasionally be much lower.}
\end{tabular}
\end{table}

\begin{table}[!tb]
\caption{Ablation for padding strategies for rotated view generation.}
\label{tab:abl-padding}
\setlength{\tabcolsep}{4.2mm}
\centering
\begin{tabular}{c|c|ccc}
\hline
\rule{0pt}{9pt}\textbf{Dataset} & \textbf{Padding} & \textbf{AP}    & \textbf{AP$_{50}$}  & \textbf{AP$_{75}$}  \\
\hline
\multirow{2}{*}{DOTA}    & Zeros   & 40.49 & 72.26 & 39.15 \\
        & Reflection   & \textbf{40.69} & \textbf{72.31} & \textbf{39.49} \\
\hline
\multirow{2}{*}{HRSC}    & Zeros    & 55.90 & 89.32 & 60.95 \\
        & Reflection    & \textbf{58.03} & \textbf{89.66} & \textbf{64.80} \\
\hline
\end{tabular}
\end{table}

\begin{table}[!tb]
\caption{Ablation with different levels of noise adding to the annotations on DOTA. }
\label{tab:abl-noise}
\setlength{\tabcolsep}{4.7mm}
\centering
\begin{tabular}{c|c|c|c}
\hline
\rule{0pt}{9pt}{$\sigma$} & \textbf{H2RBox} & \textbf{H2RBox-v2} & \textbf{Point2RBox} \\ 
\hline
0\% & \textbf{70.05} & \textbf{72.31} & \textbf{40.27}   \\
10\% & 69.19 & 71.68 & 39.60   \\
30\% & 67.39 & 71.11 & 38.42  \\
\hline
\end{tabular}
\end{table}

\begin{table}[!tb]
\caption{Ablation with fusion and scaling strategies in P2R subnet. }
\label{tab:abl-p2r}
\setlength{\tabcolsep}{4.1mm}
\centering
\begin{tabular}{c|c|c}
\hline
\rule{0pt}{9pt}{\textbf{Fusion \& Scaling}} & \textbf{Point2RBox} & \textbf{Wholly-WOOD} \\ 
\hline
$\times$ & 40.27 & 53.02   \\
\checkmark  & \textbf{51.99 (+11.72)} & \textbf{62.63 (+9.61)}   \\
\hline
\end{tabular}
\end{table}

\textbf{Padding strategies.} Compared to the performance loss of more than 10\% for H2RBox without reflection padding, Table \ref{tab:abl-padding} shows that H2RBox-v2 is less sensitive to black borders. However, reflection padding is still a better choice in the rotated view generation.

\textbf{Annotation inaccuracy.} For HBox annotations, we multiply their height and width by a noise from the uniform distribution $\left(1-\sigma, 1+\sigma \right )$. For points, we offset their coordinates by a noise from the uniform distribution $\left[-\sigma H, +\sigma H \right ]$, where $H$ is the height of objects. Table \ref{tab:abl-noise} shows that the AP$_{50}$ of H2RBox-v2 and Point2RBox drops by only 1.20\% and 1.85\% respectively when $\sigma = 30\%$, which demonstrates the robustness of the devised learning mechanisms.

\textbf{Recolor step in pattern generator.} To narrow the gap between generated synthetic patterns and real objects, we recolor the patterns based on the colors sampled around each labeled point. With this key recolor step removed (i.e. directly pasting augmented patterns like copy-paste), the AP$_{50}$ is much lower (40.27\% vs. 28.72\%) on DOTA.

\begin{figure}[t]
\centering
\includegraphics[width=0.98\linewidth]{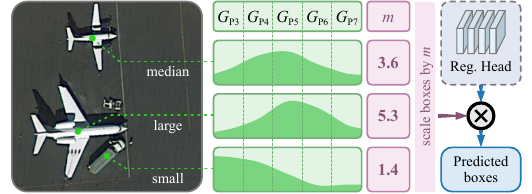}
\caption{An experiment to show that the P2R subnet can scale the output boxes based on the gating scores between FPN layers.}\label{fig:gate}
\end{figure}

\textbf{Fusion and scaling.} Figure \ref{fig:gate} shows the P2R subnet can learn the gating scores of FPN layers and scale the output boxes. Compared to merely using the P3 layer, this novel fusion mechanism improves Wholly-WOOD (Point-to-RBox) by 9.61\% (62.63\% vs. 53.02\%). 
Additionally, using our fusion and scaling strategies in the end-to-end Point2RBox \cite{yu2024point2rbox} can also boost AP$_{50}$ by 11.72\% on DOTA (see Table \ref{tab:abl-p2r}).

\textbf{Baseline detectors.} Our symmetry-aware learning approach is also effective for refine-stage and two-stage detectors. We provide Wholly-WOOD implementations based on S$^2$ANet \cite{Han2022Align} and ReDet \cite{Han2021Redet}, which achieve 72.56\% and 75.00\% in the HBox-to-RBox task on the DOTA dataset. These results are comparable to the one based on FCOS (73.48\%).

\subsection{Accuracy of Wholly-WOOD with different annotations}

We conduct quantitative experiments on three datasets that have been annotated with RBoxes: DOTA, HRSC, and FAIR1M. The RBoxes are first converted to Points/HBoxes to form the inputs of our detector. The results are displayed in Fig. \ref{fig:intro}c. The detector (i.e. Rotated FCOS \cite{Tian2019FCOS}) trained directly with RBoxes is set as the baseline for comparison (displayed as red dashed lines in the figure). 

\begin{figure}[t]
\centering
\includegraphics[width=0.98\linewidth]{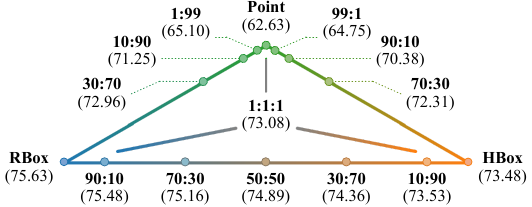}
\caption{Experiments with different Point/HBox/RBox proportions on the DOTA-v1.0 dataset. The results are formatted as ``\textbf{Proportion} (AP$_{50}$)".}\label{fig:mix}
\end{figure}

\begin{table*}[t]
\caption{Comparisons with state-of-the-art methods on DOTA-v1.0. ``RAM" denotes the RAM usage (GB) in training; ``FPS" indicates the inference speed of the trained detector; ``*" marks our conference work. AP$_{50}$ is evaluated on the test set.}
\label{tab:dotav1}
\setlength{\tabcolsep}{3.5mm}
\centering
\begin{tabular}{c|l|ccc|cc|c}
\hline
\rule{0pt}{9pt}\makebox[10mm][c]{\textbf{Anno.}} & \rule{0pt}{9pt}\makebox[47mm][c]{\textbf{Methods}} & \makebox[11mm][c]{\textbf{Sched.}} & \makebox[6mm][c]{\textbf{MS}} & \makebox[8mm][c]{\textbf{RR}} & \makebox[10mm][c]{\textbf{RAM}} & \makebox[10mm][c]{\textbf{FPS}}  & \makebox[12mm][c]{\textbf{AP$_{50}$}}  \\
\hline
\multirow{17}{*}{\textbf{RBox}} & RepPoints (2019) \cite{Yang2019Reppoints} & 1$\times$     &    &    & 3.45 & 24.5 & 68.45 \\
& RetinaNet (2017) \cite{Lin2017Focal} & 1$\times$     &    &    & 3.38 & 25.4 & 68.69 \\
& KLD (2021) \cite{Yang2021Learning} & 1$\times$     &    &    & 3.39 & 25.4 & 71.24 \\
& KFIoU (2023) \cite{yang2023kfiou} & 1$\times$     &    &    & 3.39 & 25.4 & 71.61 \\
& GWD (2021) \cite{Yang2021Rethinking} & 1$\times$     &    &    & 3.39 & 25.4 & 71.66 \\
& PSC (2023) \cite{yu2023psc}      & 1$\times$     &    &    & 4.29 & 25.4 & 71.92 \\
& SASM (2022) \cite{Hou2022Shape}  & 1$\times$     &    &    & 3.53 & 24.4 & 72.30  \\
& R$^3$Det (2021) \cite{Yang2021R3Det} & 1$\times$     &    &    & 3.54 & 20.0 & 73.12 \\
& CFA (2021) \cite{Guo2021Beyond}  & 1$\times$     &    &    & 3.45 & 24.5 & 73.84 \\
& Oriented RepPoints (2022) \cite{li2022oriented} & 1$\times$     &    &    & 3.45 & 24.5 & 75.26 \\
& S$^2$A-Net (2022) \cite{Han2022Align} & 1$\times$     &    &    & 3.14 & 23.3 & 75.81 \\
& FCOS (2019) \cite{Tian2019FCOS} (baseline) & 1$\times$     &    &    & 4.18 & 29.5 & 72.44 \\
& FCOS (2019) \cite{Tian2019FCOS} (baseline) & 3$\times$     &    & \checkmark & 4.18 & 29.5 & 74.75 \\
& FCOS (2019) \cite{Tian2019FCOS} (baseline) & 1$\times$     & \checkmark & \checkmark & 4.18 & 29.5 & 77.68 \\
\rowcolor{gray!20} \cellcolor{white} & Wholly-WOOD (ours, FCOS-based) & 1$\times$     &    &    & 6.67 & 29.1 & 75.63 \\
\rowcolor{gray!20} \cellcolor{white} & Wholly-WOOD (ours, FCOS-based) & 3$\times$     &    & \checkmark & 6.67 & 29.1 & 76.39 \\
\rowcolor{gray!20} \cellcolor{white} & Wholly-WOOD (ours, FCOS-based) & 1$\times$     & \checkmark & \checkmark & 6.67 & 29.1 & \textbf{79.40} \\
\hline
\multirow{15}{*}{\textbf{HBox}} & BoxInst-RBox (2021) \cite{tian2021boxinst}$^1$ & 1$\times$     &    &    & 20.34  & 2.7  & 53.59 \\
& BoxLevelSet-RBox (2022) \cite{li2022box}$^1$ & 1$\times$     &    &    & $>$24$^2$  & 4.7  & 56.44 \\
& SAM-ViT-B-RBox (2023) \cite{kirillov2023segany}$^{1,3}$ & 1$\times$     &    &    & -  & 1.7 & 63.94 \\
& EIE-Det (2024) \cite{wang2024explicit} & 1$\times$ & & & - & 29.1 & 70.08 \\
& EIE-Det (2024) \cite{wang2024explicit} & 1$\times$ & \checkmark & & - & 29.1 & 75.74 \\
& H2RBox (2023) \cite{yang2023h2rbox}$^*$ & 1$\times$     &    &    & 6.55 & 29.1 & 67.82 \\
& H2RBox (2023) \cite{yang2023h2rbox}$^*$ & 1$\times$     &  \checkmark  &    & 6.55 & 29.1 & 74.40 \\
& H2RBox-v2 (2023) \cite{yu2023h2rboxv2}$^*$              & 1$\times$     &    &  & 10.10 & 29.1 & 72.31 \\
& H2RBox-v2 (2023) \cite{yu2023h2rboxv2}$^*$              & 3$\times$     &    & \checkmark & 10.10 & 29.1 & 74.29 \\
& H2RBox-v2 (2023) \cite{yu2023h2rboxv2}$^*$             & 1$\times$     & \checkmark & \checkmark & 10.10 & 29.1 & 78.25 \\
& AFWS (2024) \cite{lu2024afws} & 1$\times$ & & & - & 29.1 & 72.55 \\
& AFWS (2024) \cite{lu2024afws} & 1$\times$ & \checkmark & & - & 29.1 & 78.13 \\

\rowcolor{gray!20} \cellcolor{white} & Wholly-WOOD (ours, FCOS-based) & 1$\times$     &    &    & 6.67 & 29.1 & 73.48 \\
\rowcolor{gray!20} \cellcolor{white} & Wholly-WOOD (ours, FCOS-based) & 3$\times$     &    & \checkmark & 6.67 & 29.1 & 74.94 \\
\rowcolor{gray!20} \cellcolor{white} & Wholly-WOOD (ours, FCOS-based) & 1$\times$     & \checkmark & \checkmark & 6.67 & 29.1 & \textbf{78.24} \\
\hline
\multirow{12}{*}{\textbf{Point}} & Point2Mask-RBox (2023) \cite{li2023point2mask}$^1$ & 1$\times$     &    &    & 16.97  & 9.5  & 9.72 \\
& P2BNet+H2RBox (2023) \cite{chen2022pointtobox,yang2023h2rbox} & 1$\times$     &    &    & $>$24$^4$  & 29.1 & 19.63 \\
& P2BNet+H2RBox-v2 (2023) \cite{chen2022pointtobox,yu2023h2rboxv2} & 1$\times$     &    &    & $>$24$^4$  & 29.1 & 21.87 \\
& P2RBox (SAM-based) (2023) \cite{cao2023p2rbox}$^3$            & 1$\times$     &    &    & -  & 29.1 & 58.40 \\
& PointOBB (2024) \cite{luo2024pointobb}            & 1$\times$     &    &    & $>$24$^4$  & 29.1 & 30.08 \\
& Point2RBox (2024) \cite{yu2024point2rbox}$^*$            & 1$\times$     &    &    & 7.52  & 29.1 & 40.27 \\
& PointOBB-v2 (2025) \cite{ren2025pointobbv2}            & 1$\times$     &    &    & 5.99  & 29.1 & 41.68 \\
& PointOBB-v3 (2025) \cite{zhang2025pointobbv3}            & 1$\times$     &    &    & $>$24$^4$  & 29.1 & 49.24 \\
& Point2RBox-v2 (2025) \cite{yu2025point2rboxv2}            & 1$\times$     &    &    & 6.30  & 29.1 & 62.61 \\
\rowcolor{gray!20} \cellcolor{white} & Wholly-WOOD (ours, FCOS-based)             & 1$\times$     &    &    & 6.67  & 29.1 & 62.63 \\
\rowcolor{gray!20} \cellcolor{white} & Wholly-WOOD (ours, FCOS-based)             & 3$\times$     &    & \checkmark & 6.67  & 29.1 & 63.10 \\
\rowcolor{gray!20} \cellcolor{white} & Wholly-WOOD (ours, FCOS-based)             & 1$\times$     & \checkmark & \checkmark & 6.67  & 29.1 & \textbf{65.10} \\
\hline
\specialrule{0pt}{2pt}{0pt}
\multicolumn{3}{l}{$^1$Minimum rectangle operation is performed on Mask to obtain RBox.} &
\multicolumn{5}{l}{$^2$Evaluated on NVIDIA V100 GPU due to excessive RAM usage.} \\
\multicolumn{3}{l}{$^3$Using the SAM model \cite{kirillov2023segany} pre-trained on massive additional data.} &
\multicolumn{5}{l}{$^4$Depending on instance count, capped at 100 per image for 24\,GB.} \\
\end{tabular}
\end{table*}

\textbf{RBox-supervision setting.} While Wholly-WOOD aims primarily at weakly-supervised learning, it also incorporates support for RBox annotations. Notably, when utilizing RBox supervision, the accuracy of Wholly-WOOD (represented by the blue bars in Fig. \ref{fig:intro}c) also exhibits a slight improvement over the FCOS baseline, largely attributed to the self-supervision module within the training pipeline.

\textbf{HBox-to-RBox setting.} Note when trained with HBoxes, our detector achieves a performance close to that of the RBox-trained counterpart. In concrete terms, the performance of our method is 1.04\% (w/o MS and RR) and 0.19\% (w/ RR) higher than the RBox-supervised FCOS baseline on DOTA-v1.0. When MS and RR are both applied, it outperforms RBox-supervised FCOS by 0.56\% (78.24\% vs. 77.68\%). On the more challenging DOTA-v1.5/2.0 datasets, the results present a similar trend, whereas on the FAIR1M dataset, Wholly-WOOD performs superior to the RBox-supervised FCOS by 1.93\% (43.18\% vs. 41.25\%). Overall, Wholly-WOOD, merely using HBox annotations, outperforms RBox-supervised baseline by 0.98\% average over the five datasets (w/o MS and RR), proving that our weakly-supervised learning paradigm can achieve performance on a par with the fully-supervised one upon the same detector in HBox-to-RBox setting.

\begin{figure*}[t]
\centering
\includegraphics[width=0.935\textwidth]{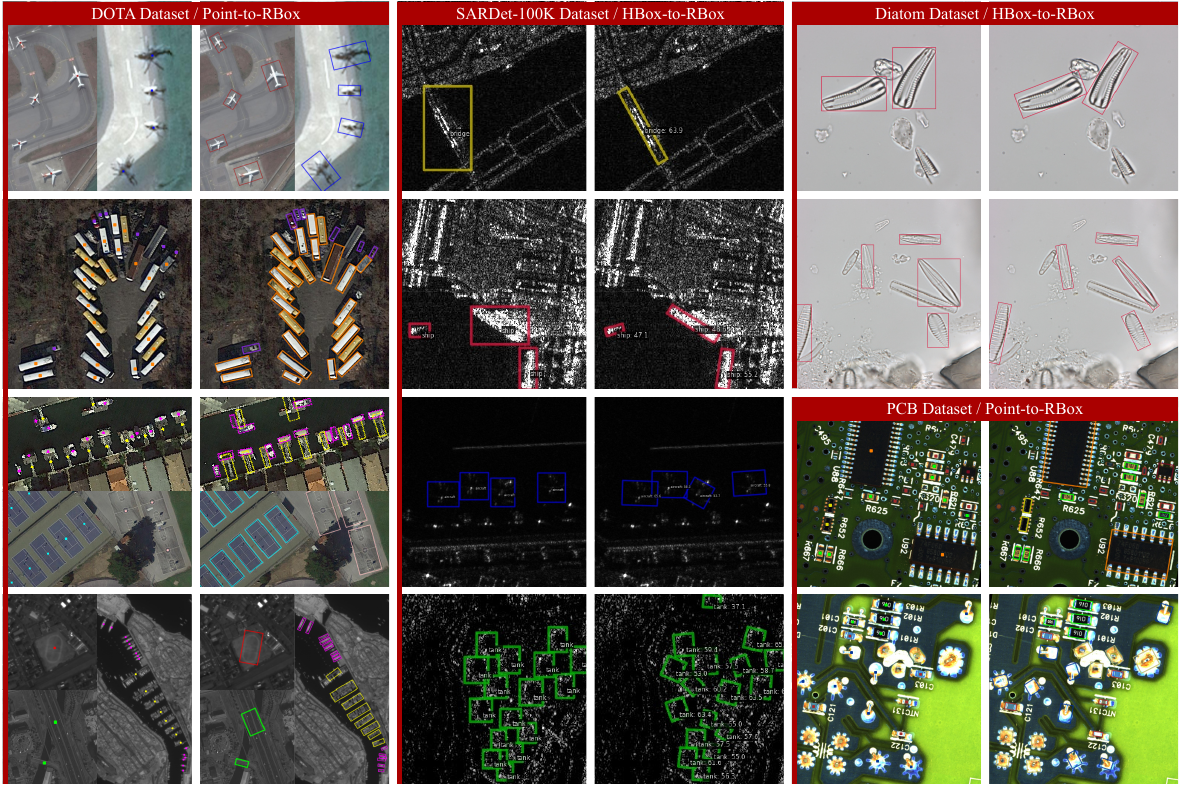}
\caption{Experimental results of our Wholly-WOOD. From left to right: \textbf{1)} Point-to-RBox conversion on the DOTA dataset; \textbf{2)} HBox-to-RBox conversion on the HBox-annotated SARDet-100K dataset; \textbf{3)} Applicability to other scenarios including diatom detection and PCB component detection.}\label{fig:expa}
\end{figure*}

\textbf{Point-to-RBox setting.} When trained with more coarse-grained point annotations, our method gives an AP$_{50}$ performance 9.81\% lower than the RBox-trained baseline (62.63\% vs. 72.44\%) on DOTA-v1.0. Although the boxes are not as accurate as HBox/RBox-supervised settings, they are quite sufficient for many applications (see the visualization in Fig. \ref{fig:expa}). Since DOTA-v1.0 contains 15 different classes of remote-sensing objects, such results also demonstrate the broad applicability of our approach. On the HRSC dataset for ship detection, the gap is only 1.69\% (87.30\% vs. 88.99\%). Accuracy for each category of DOTA-v1.0 (MSRR) in Fig. \ref{fig:intro}b reveals that our Point-to-RBox conversion achieves near-optimal accuracy for numerous categories. However, there remains a discernible gap for categories characterized by less distinct boundaries (i.e. Bridge, Soccer-Ball-Field, and Harbor).

\textbf{Combination of diverse labels.} Figure \ref{fig:mix} shows the detection performance of our detector across different combinations of two annotation formats. We demonstrate that incorporating a small proportion of RBoxes/HBoxes in the Point setting can notably enhance the accuracy. When training on the DOTA-v1.0 dataset with a mix of 70\% Points and 30\% HBoxes, we achieve an AP$_{50}$ accuracy on the test set of 72.31\%, approaching that of RBox-supervised FCOS.

\subsection{Comparisons with state-of-the-art methods}

\textbf{DOTA-v1.0.} Table \ref{tab:dotav1} demonstrates that in the HBox-to-RBox setting, the performance gap between our method and the RBox-supervised FCOS baseline is minimal. In Point-to-RBox conversion, while a 9.81\% gap persists, Wholly-WOOD achieves competitive accuracy compared to state-of-the-art methods (e.g. PointOBB \cite{luo2024pointobb}, 62.63\% vs. 30.08\%).

\begin{table*}[!tb]
\caption{Accuracy comparisons on the DOTA-v1.0/1.5/2.0, HRSC, FAIR1M, and STAR datasets.}
\label{tab:otherds}
\setlength{\tabcolsep}{3.2mm}
\centering
\begin{tabular}{l|c|c|c|c|c|c}
\hline
\makebox[30mm][c]{\textbf{Method}} & \makebox[19mm][c]{\textbf{DOTA-v1.0}} & \makebox[19mm][c]{\textbf{DOTA-v1.5}} & \makebox[19mm][c]{\textbf{DOTA-v2.0}} & \makebox[13mm][c]{\textbf{HRSC}} & \makebox[13mm][c]{\textbf{FAIR1M}} \rule{0pt}{9pt} & \makebox[13mm][c]{\textbf{STAR}} \rule{0pt}{9pt} \\
\hline
RetinaNet (2017) \cite{Lin2017Focal} & 68.69 & 60.57        & 47.00        & 84.49   & 37.67   & 21.80  \\
GWD (2021) \cite{Yang2021Rethinking} & 71.66 & 63.27        & 48.87        & 86.67   & 39.11   & 25.30  \\
S$^2$A-Net (2022) \cite{Han2022Align} & \textbf{75.81} & \textbf{66.53}        & \textbf{52.39}        & \textbf{90.10}  & \textbf{42.44}   & 27.30 \\
FCOS (2019) \cite{Tian2019FCOS} & 72.44 & 64.53        & 51.77        & 88.99  & 41.25   & \textbf{28.10} \\
\hline
Sun et al. (2021) \cite{sun2021oriented}$^1$ & 38.60 & - & - & - & - & - \\
KCR (2023) \cite{zhu2023knowledge}$^2$ & - & -        & -        &  79.10  & -  & -  \\
H2RBox (2023) \cite{yang2023h2rbox} & 70.05 & 61.70        & 48.68        &  7.03  & 35.94  & 17.20    \\
H2RBox-v2 (2023) \cite{yu2023h2rboxv2} & 72.31 & 64.76        & 50.33        & 89.66   & 42.27  & 27.30    \\
AFWS (2024) \cite{lu2024afws} & 72.55 & \textbf{65.92} & 51.73 & - & 41.80 & - \\
Wholly-WOOD (ours) & \textbf{73.48} & 65.27        & \textbf{52.13}        & \textbf{89.80}   & \textbf{43.18}  & \textbf{27.50}    \\
\hline
P2RBox (2024) \cite{cao2023p2rbox}$^3$ & 58.40 & -        & -        & -   & -  & -    \\
PointOBB (2024) \cite{luo2024pointobb} & 30.08 & 10.66        & 5.53        & -   & 11.19  & -    \\
Point2RBox (2024) \cite{yu2024point2rbox} & 40.27 & 30.51        & 23.43        & 79.40   & 20.03  & -    \\
PointOBB-v2 (2025) \cite{ren2025pointobbv2} & 41.68 & 30.59        & 20.64        & -   & 13.36  & -    \\
PointOBB-v3 (2025) \cite{zhang2025pointobbv3} & 49.24 & 33.79        & 23.52        & -   & 18.35  & -    \\
Wholly-WOOD (ours) & \textbf{62.63} & \textbf{52.78}        & \textbf{38.16}        & \textbf{87.30}   & \textbf{32.83}  & \textbf{-}    \\
\hline
\specialrule{0pt}{2pt}{0pt}
\multicolumn{3}{l}{$^1$Sparse annotation for horizontal/vertical objects. }&
\multicolumn{4}{l}{$^2$Transfer learning from DOTA (RBox) to HRSC (HBox). }\\
\multicolumn{7}{l}{$^3$Using the SAM model \cite{kirillov2023segany} pre-trained on massive additional data.}
\end{tabular}
\end{table*}

Methods potentially applicable to Point/HBox-to-RBox task setting in a cascade manner are also compared in Table \ref{tab:dotav1}. \textbf{1)} Point/HBox-to-Mask-to-RBox. Weakly-supervised methods (e.g. BoxInst \cite{tian2021boxinst}, BoxLevelSet \cite{li2022box}, and Point2Mask \cite{li2023point2mask}) can be applied to oriented detection tasks since the segmentation mask can be converted to RBox by finding the minimum circumscribed rectangle. \textbf{2)} Point-to-HBox-to-RBox. P2BNet \cite{chen2022pointtobox} samples boxes of different sizes around the labeled point and classify them through Multiple Instance Learning (MIL) to achieve Point-to-HBox. RBoxes can be obtained by using a subsequent HBox-to-RBox stage. 

Table \ref{tab:dotav1} shows that our method outperforms these cascade solutions in both accuracy and speed. Taking BoxLevelSet-RBox \cite{li2022box} as an example, Wholly-WOOD (HBox-to-RBox) gives an accuracy of 17.04\% higher and a speed 7$\times$ faster by avoiding the time-consuming post-processing (i.e. minimum circumscribed rectangle operation). In particular, the foundation model for segmentation SAM \cite{kirillov2023segany} has shown strong zero-shot capabilities by training on the largest segmentation dataset to date. Benefiting from its powerful zero-shot capability, SAM-ViT-B-RBox achieved an accuracy of 63.94\% on HBox-to-RBox conversion, while P2RBox \cite{cao2023p2rbox} attained 58.40\% on Point-to-RBox. When compared to SAM-based approaches, Wholly-WOOD still delivers superior accuracy in both HBox and Point settings.

\begin{figure*}[t]
\centering
\includegraphics[width=0.935\textwidth]{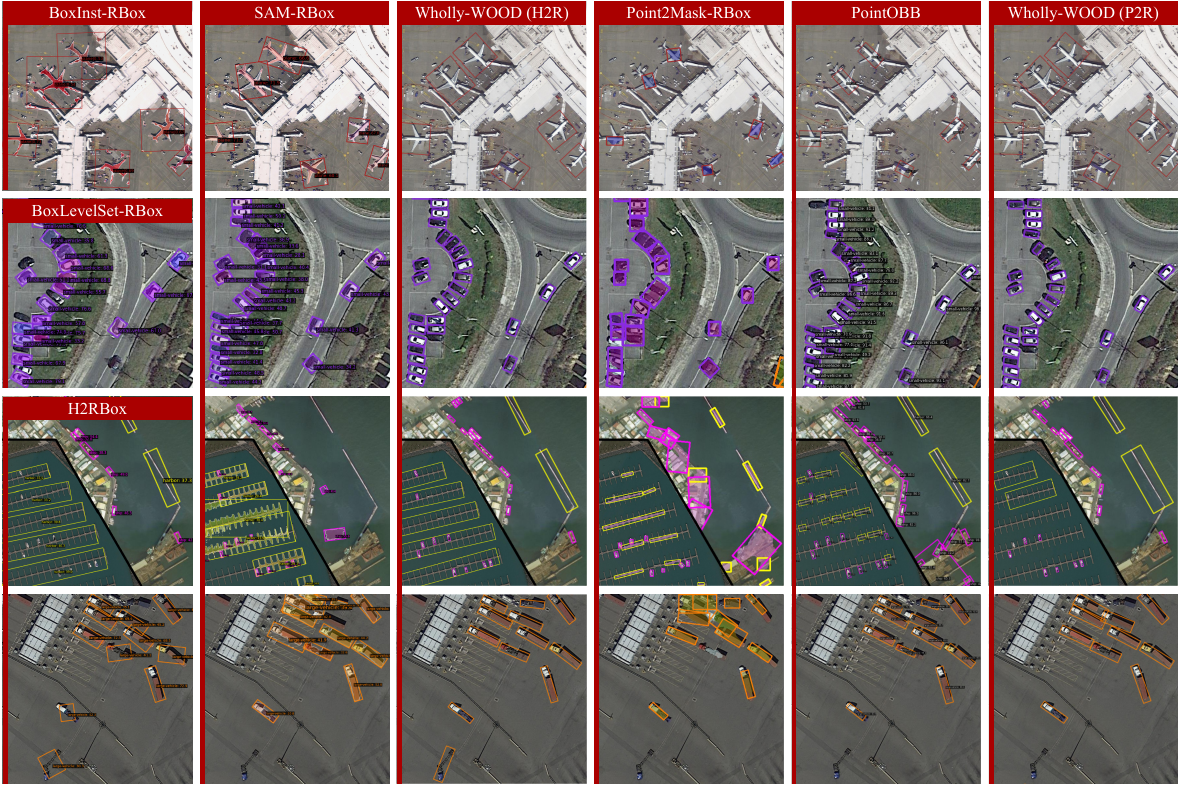}
\caption{Visualization to compare the state-of-the-art approaches and our Wholly-WOOD. The first three columns are HBox-to-RBox methods and the others are Point-to-RBox ones. For segmentation methods, the suffix ``-RBox" indicates using minimum rectangle operation on Mask to obtain RBox.}
\label{fig:expb}
\end{figure*}

In comparison with our conference versions, the proposed Wholly-WOOD detector also demonstrates considerable improvements. While the accuracy of Wholly-WOOD in HBox-to-RBox conversion is similar to H2RBox-v2 \cite{yu2023h2rboxv2}, the RAM usage is further reduced to 6.67\,GB with our enhanced architecture. In terms of Point-to-RBox, the accuracy is significantly improved by 22.36\% (62.63\% vs. 40.27\%) with lower RAM usage compared to Point2RBox \cite{yu2024point2rbox}.

\textbf{DOTA-v1.5/2.0.} As extended versions of DOTA-v1.0, these two datasets are more challenging, while the results present a similar trend. Still, Wholly-WOOD shows an HBox-to-RBox conversion accuracy slightly higher than its RBox-trained FCOS counterpart (65.27\% vs. 64.53\% on DOTA-v1.5 and 52.13\% vs. 51.77\% on DOTA-v2.0, see Table \ref{tab:otherds}).

\textbf{HRSC.} Our previous work H2RBox \cite{yang2023h2rbox} can hardly learn angle information from small datasets like HRSC, resulting in deficient performance. Contrarily, H2RBox-v2 and Wholly-WOOD give an HBox-to-RBox performance comparable to fully-supervised methods. Compared to KCR \cite{zhu2023knowledge} that uses transfer learning from RBox-supervised DOTA to HBox-supervised HRSC, Wholly-WOOD, merely using HBox, outperforms KCR by 10.70\% (89.80\% vs. 79.10\%).

\textbf{FAIR1M.} This dataset contains a large number of planes, vehicles, and courts, which are more perfectly symmetric than objects like harbors in DOTA. This may explain the observation that symmetry-aware learning (H2RBox-v2 and Wholly-WOOD), outperforms H2RBox by a more considerable margin. In this case, Wholly-WOOD performs superior to the RBox-supervised FCOS by 1.93\% (43.18\% vs. 41.25\%). 

\textbf{STAR.} Facing 48 fine-grained categories of diverse spatial resolutions, Wholly-WOOD still gives a comparable accuracy close to RBox-supervised FCOS (27.50\% vs. 28.10\%), proving the wide applicability of our method.

Accuracy and RAM usage aside, Wholly-WOOD presents an added advantage by unifying various weak-supervision tasks. Integrating Point/HBox/RBox annotations, or their combination, into a unified pipeline, our detector offers users a more convenient and versatile solution. Figure \ref{fig:expb} visualizes the comparisons among the state-of-the-art approaches.

\subsection{Experiments on real Point/HBox labeled datasets}

The above experiments are based on RBox-labeled datasets by degrading the RBoxes to HBoxes/Points for training. To validate the detection performance of Wholly-WOOD in real label reduction scenarios, SARDet-100K \cite{li2024sardet100k}, a dataset with only HBox annotations is used as the input of our detector. Although there is no ground truth for quantitative analysis, the visualization results in Fig. \ref{fig:expa} show that our detector successfully obtains quite accurate RBox annotations.

Furthermore, experiments on diatom images\footnote{\url{https://doi.org/10.34740/kaggle/ds/1187591}.} and PCB images\footnote{\url{https://doi.org/10.34740/kaggle/ds/5060183}.} are carried out to validate the applicability of our approach in scenarios other than remote sensing. Figure \ref{fig:expa} demonstrates that our detector can also reduce the annotation in other oriented object detection tasks.

\subsection{Further discussion}\label{sec:diss}

While horizontal ground-truths are well-established in object detection, retrieving rotated ones is laborious, requiring highly trained experts and often resulting in imprecision. This emphasizes the need for weakly-supervised deep learning approaches that do not rely on rotated annotations but instead leverage annotations that are easier and faster to obtain. 

How much annotation task can be reduced by utilizing HBox/Point supervision? An instinct concept is that compared to RBoxes, HBox annotations reduce the workload from three clicks to two, whereas Point annotations further streamline this process to just one click. However, acquiring a horizontal box annotation is straightforward, particularly with the assistance of a cross-line on the screen for accurate alignment. Despite appearing to require just one more parameter, the process of obtaining a rotated box can be more time-consuming than expected due to its five degrees of freedom. 

Typically, there are two ways to annotate rotated boxes:
\textbf{1)} Draw a polygon shape with four clicks around the object of interest and then convert it into a rotated box. 
\textbf{2)} Draw a horizontal bounding box around the object, then rotate it to align with the object's orientation, and finally adjust the width and height again. To quantify the time required for different annotation formats, we conduct a user study wherein experienced annotators are tasked with annotating an image from the DOTA-v1.0 dataset \cite{Xia2018DOTA} using the second way. The results indicate that, on average, it takes 1.07 seconds for Point annotation, 2.23 seconds for HBox annotation, and 3.69 seconds for RBox annotation for a single instance.

It can be inferred from these results that utilizing Wholly-WOOD for HBox supervision can lead to a reduction in annotation time by 40\% while maintaining comparable detection accuracy. Alternatively, employing the Point-to-RBox setting can achieve a time reduction of 71\% if a slight accuracy trade-off is acceptable (the evaluated AP$_{50}$ loss is 9.81\% and 1.69\% on the DOTA-v1.0 and HRSC datasets).

\section{Conclusion}\label{sec:con}

In this work, we have introduced Wholly-WOOD, a unified weakly-supervised detector aimed at wholly leveraging diversified-quality labels for oriented object detection, demonstrating its effectiveness in remote sensing and beyond. 

Through extensive experiments, we make the following observations: 
\textbf{1)} Our approach enables the unification of data with various annotation formats, offering a more convenient and versatile solution with accuracy surpassing other state-of-the-art alternatives. 
\textbf{2)} The use of Wholly-WOOD for HBox-to-RBox learning leads to a reduction in annotation time by 40\% while maintaining comparable detection accuracy. 
\textbf{3)} Employing Point-to-RBox achieves a time reduction of 71\% with a marginal accuracy loss of 9.81\% and 1.69\% on DOTA-v1.0 and HRSC, respectively.
\textbf{4)} Using diversified-quality labels could be a good alternative to balance the annotation and accuracy. When RBox:HBox:Point = 1:1:1, the accuracy on DOTA-v1.0 reaches 73.08\%, quite close to the FCOS detector fully supervised by RBoxes. 

Wholly-WOOD illustrates the effectiveness of Point/HBox weak supervision, delivering detection performance similar to its RBox-supervised counterpart, making it an unprecedented alternative for processing annotations of various formats in oriented object detection tasks. We believe this research can help alleviate the burden of costly manual annotation, freeing individuals from labor-intensive labeling tasks.

\bibliographystyle{IEEEtran}
\bibliography{egbib}

\end{document}